\documentclass[conference]{IEEEtran}
\IEEEoverridecommandlockouts
\usepackage{times}

\usepackage[numbers]{natbib}
\usepackage{multicol}
\usepackage{graphicx}
\usepackage{booktabs}
\usepackage{subcaption}
\usepackage{amsmath}
\usepackage{tikz}
\usetikzlibrary{shapes.geometric, arrows.meta, positioning}
\usepackage[bookmarks=true]{hyperref}
\hypersetup{colorlinks=true,linkcolor=black,citecolor=black,urlcolor=blue,pdfauthor={Abhishek Seetharamu, Badrikanath Praharaj, Sreeram M.V.},pdftitle={Subterra: An Open-Source Validated Benchmark for GNSS-Denied Tunnel Navigation}}
\begin{document}

\title{Subterra: An Open-Source Validated Benchmark for GNSS-Denied Tunnel Navigation}


\author{
\IEEEauthorblockN{Abhishek Seetharamu\textsuperscript{*}}
\IEEEauthorblockA{\textit{BuildMachineLabs}\\
\href{mailto:abhishekss6363@gmail.com}{\textcolor{blue}{mail}}}
\and
\IEEEauthorblockN{Badrikanath Praharaj\textsuperscript{*}}
\IEEEauthorblockA{\textit{BuildMachineLabs}\\
\href{mailto:badrikanath.praharaj@gmail.com}{\textcolor{blue}{mail}}}
\and
\IEEEauthorblockN{Sreeram M.V.\textsuperscript{*}}
\IEEEauthorblockA{\textit{BuildMachineLabs}\\
\href{mailto:mvsreeramblr@gmail.com}{\textcolor{blue}{mail}}}
\thanks{\textsuperscript{*}These authors contributed equally to this work.}
}

\maketitle

\begin{abstract}
Underground and subway tunnels are among the hardest settings for autonomous
navigation: they are Global Navigation Satellite System (GNSS) denied because no satellite signal penetrates the
surrounding rock, and geometrically degenerate---long, featureless, near-uniform
cross-sections offer almost no landmarks, so onboard state estimation drifts
along the tunnel axis. We present \textbf{Subterra}, an open-source benchmark for
GNSS-denied tunnel navigation whose difficulty is inherited rather than designed:
its geometry is reconstructed directly from real LiDAR in the GEODE dataset and
released as a MuJoCo environment with a Gymnasium-compatible POMDP interface. The
reconstructed bore is a $351$\,m swept tunnel; observations come from a noisy
planar LiDAR and a drifting odometry model, and a Unitree~B2 quadruped is
provided as a locomotion embodiment. A reproducible $18$-category, $199$-metric
suite scores the reconstruction at a global $\mathbf{97.1\%}$ fidelity, with a
$\mathbf{2.46}$\,\textbf{cm} synthetic-LiDAR surface error against the real
reference. As a usability check, a PPO agent trained end-to-end in Subterra
reaches a mean rolling coverage of $\mathbf{90.02\%} \pm 0.01\%$ on all five
seeds, confirming that the benchmark is hard yet learnable.
\end{abstract}

\IEEEpeerreviewmaketitle

\section{Introduction}
Autonomous robots deployed in tunnels and mines face a compounding perception
failure: no satellite signal penetrates the surrounding rock, and long
near-uniform cross-sections offer no landmarks for state estimation, so even
LiDAR odometry drifts irreversibly along the tunnel axis.
Navigation therefore reduces to a partially observable Markov decision process
(POMDP)~\cite{kaelbling1998pomdp}, the regime in which GNSS-denied odometry
drift is the dominant failure mode~\cite{zhang2016degeneracy}.

Progress is bottlenecked by \emph{evaluation}: synthetic benchmarks author
their own difficulty, which a reviewer can dispute. \textbf{Subterra}
reconstructs its geometry from real GEODE LiDAR~\cite{geode2024}, exports it
to MuJoCo~\cite{todorov2012mujoco} behind a Gymnasium interface~\cite{towers2024gymnasium},
and pairs it with a noisy-LiDAR-plus-drifting-odometry POMDP.
Our contributions are:
\begin{itemize}
\item[\textbf{C1}] Subterra, an open-source
GNSS-denied tunnel-navigation POMDP reconstructed from real GEODE LiDAR and
released as a MuJoCo scene and a Gymnasium environment---furnished from the data
with railway track structure and driven by a noisy-LiDAR-plus-drifting-odometry
sensor model.
\item[\textbf{C2}] A reproducible suite of $18$
categories / $199$ logged metrics that establishes that the reconstruction
faithfully reproduces the real bore (global $97.1\%$; $2.46$\,cm surface error).
\end{itemize}
Reinforcement learning is \emph{not} a contribution of this paper: we train a PPO
agent only as a usability check that standard algorithms can learn to navigate
the released environment (Sec.~\ref{sec:results}).

\begin{figure}[t]
\centering
\setlength{\tabcolsep}{1.5pt}
\begin{tabular}{cc}
\includegraphics[width=0.485\columnwidth,height=0.355\columnwidth]{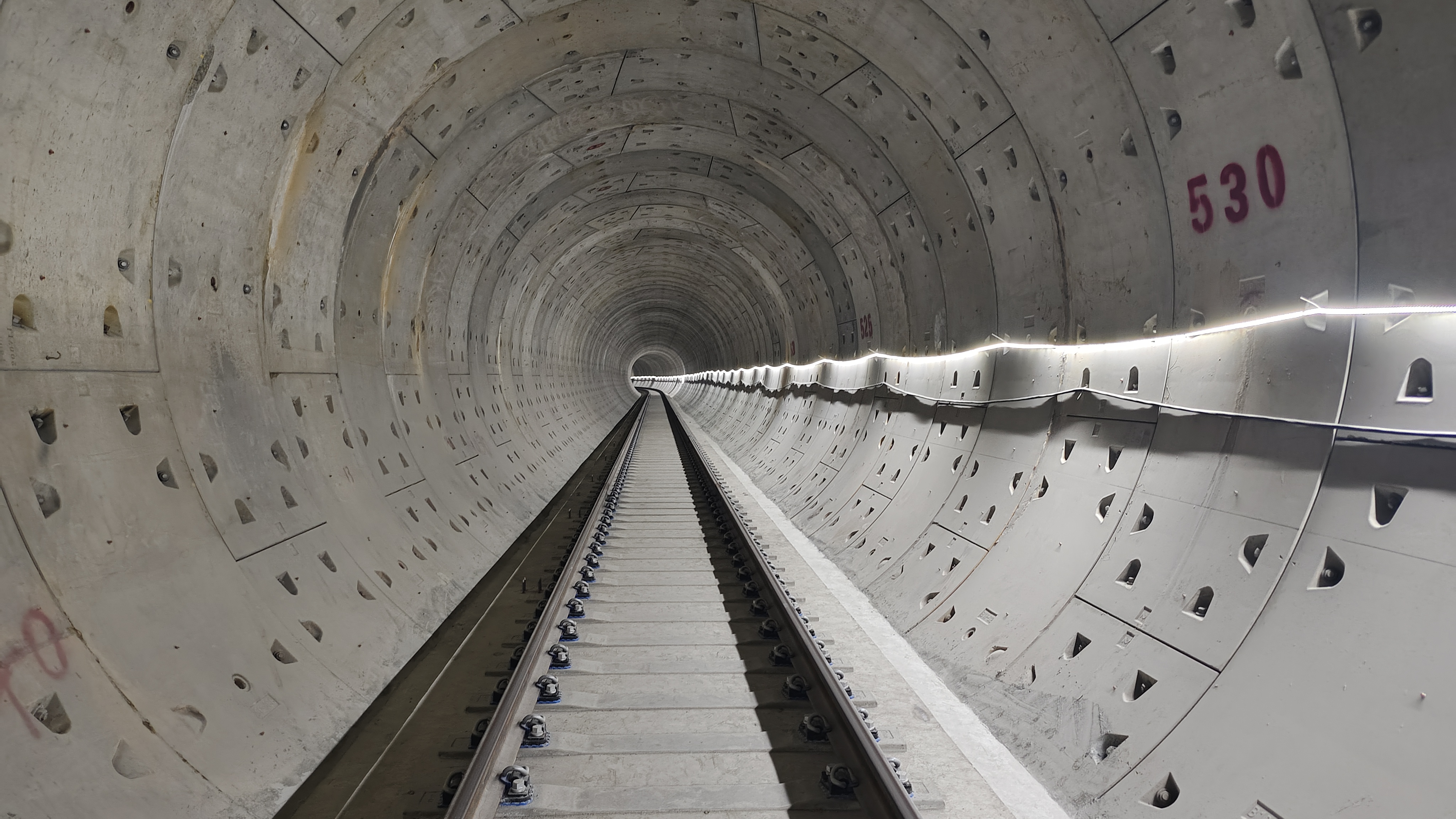} &
\includegraphics[width=0.485\columnwidth, height=0.355\columnwidth]{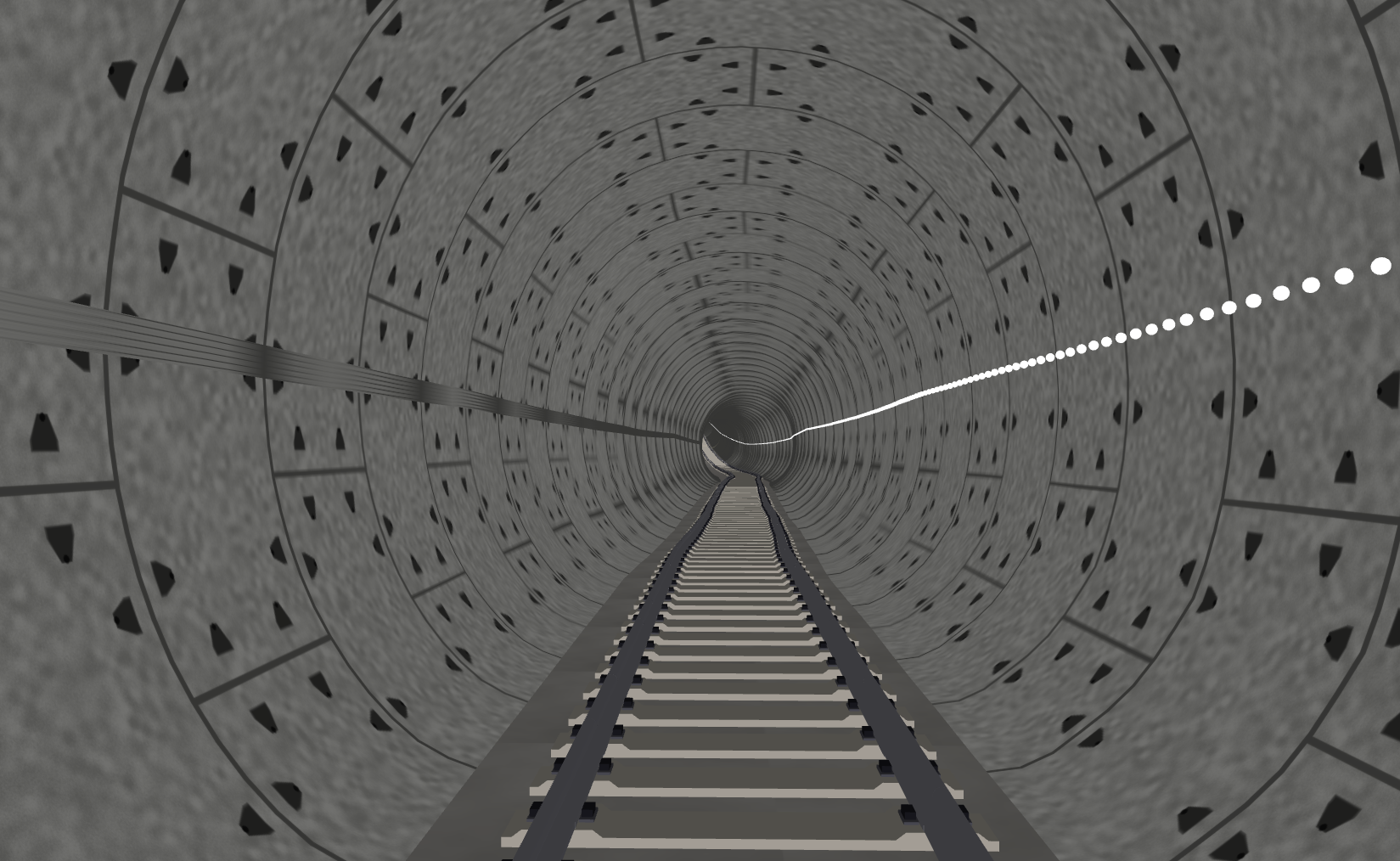} \\[1pt]
\end{tabular}

\caption{\textbf{Subterra: real vs.\ reconstructed.} Real GEODE tunnel
frame (left) and MuJoCo bore environment (right).}
\label{fig:env}
\end{figure}

\section{The Subterra Environment}
\label{sec:env}
\textbf{Real2Sim reconstruction.} We use an underground tunnel sequence from
GEODE~\cite{geode2024}, a real Livox BMI1088 LiDAR, MTi-30 IMU and stereo dataset built to stress SLAM in
geometrically degenerate environments. From the accumulated point cloud data Subterra
extracts a $351$\,m centerline and sweeps a D-section bore
($\approx\!5.65\times6.36$\,m) along it, exported as a MuJoCo
scene~\cite{todorov2012mujoco} (Fig.~\ref{fig:env}).

\textbf{Physical tunnel structure.} The bore is lined like a real bored tunnel:
featureless segmental concrete walls that offer no axial landmark. The wall
pattern is a grid of recessed holes $\approx\!13$\,cm deep, each joined to the
next cell by a connecting nail---the segment-jointing detail visible in the
source environment. Subterra also reconstructs $586$ concrete monoblock sleepers
at $0.6$\,m spacing, paired rails at the standard $1.435$\,m gauge on raised
pads, Pandrol-style fasteners at each rail seat.
The gauge is fixed by the track standard and each sleeper's fit is validated
against the local bore half-width, so every dimension is set by the track
standard or the LiDAR bore rather than hand-tuned.

\textbf{POMDP formulation.} The hidden state is the robot's pose along a deep,
GNSS-denied tunnel; the agent acts with continuous velocity commands
$a=[v,\omega]$, $v\!\in\![0,2]\,\mathrm{m/s}$,
$\omega\!\in\![-1.5,1.5]\,\mathrm{rad/s}$. A $16$-beam planar LiDAR ($30$\,m
range, $\pm5$\,cm Gaussian noise) pins lateral offset and heading but is
\emph{ambiguous along the axis}. Proprioceptive odometry carries per-step noise
\emph{and} a per-episode systematic bias, so dead-reckoning \emph{drifts}---the
core failure mode of GNSS-denied operation. The Gymnasium reset/step interface
exposes an $18$-D observation (LiDAR $+$ drifting odometry, no absolute pose), and
a Unitree~B2 quadruped is provided as a $3$-D locomotion embodiment.
\textbf{Reward.}\cite{abhishek2026minexplore} At each step the agent receives
\begin{equation}
r_t = \underbrace{r_{\text{term}}}_{\text{sparse terminal}}
      + \underbrace{g\,\frac{\Delta s_t}{L}}_{\text{coverage}}
      - \underbrace{c_d\,\frac{|d_t|}{w_t}}_{\text{lateral offset}}
      - \underbrace{c_s\big(|\Delta v_t| + |\Delta \omega_t|\big)}_{\text{control jerk}},
\label{eq:reward}
\end{equation}
where $r_{\text{term}}$ is a sparse terminal reward of $+1$ for success (reaching $90\%$ tunnel coverage without wall contact) or $0$ otherwise, granted once at episode end;
$\frac{\Delta s_t}{L}$ is the coverage reward, where $\Delta s_t$ is the newly covered arc-length beyond the furthest point reached so far (a ratchet mechanism preventing reward for re-covering ground) and $L$ is the total tunnel length, scaled by gain $g$;
$\frac{|d_t|}{w_t}$ is the lateral offset penalty, where $d_t$ is the deviation from the centerline and $w_t$ is the local half-width, penalizing off-centre travel with weight $c_d$; and
$|\Delta v_t| + |\Delta \omega_t|$ is the control jerk penalty penalizing changes in linear and angular velocity with weight $c_s$.

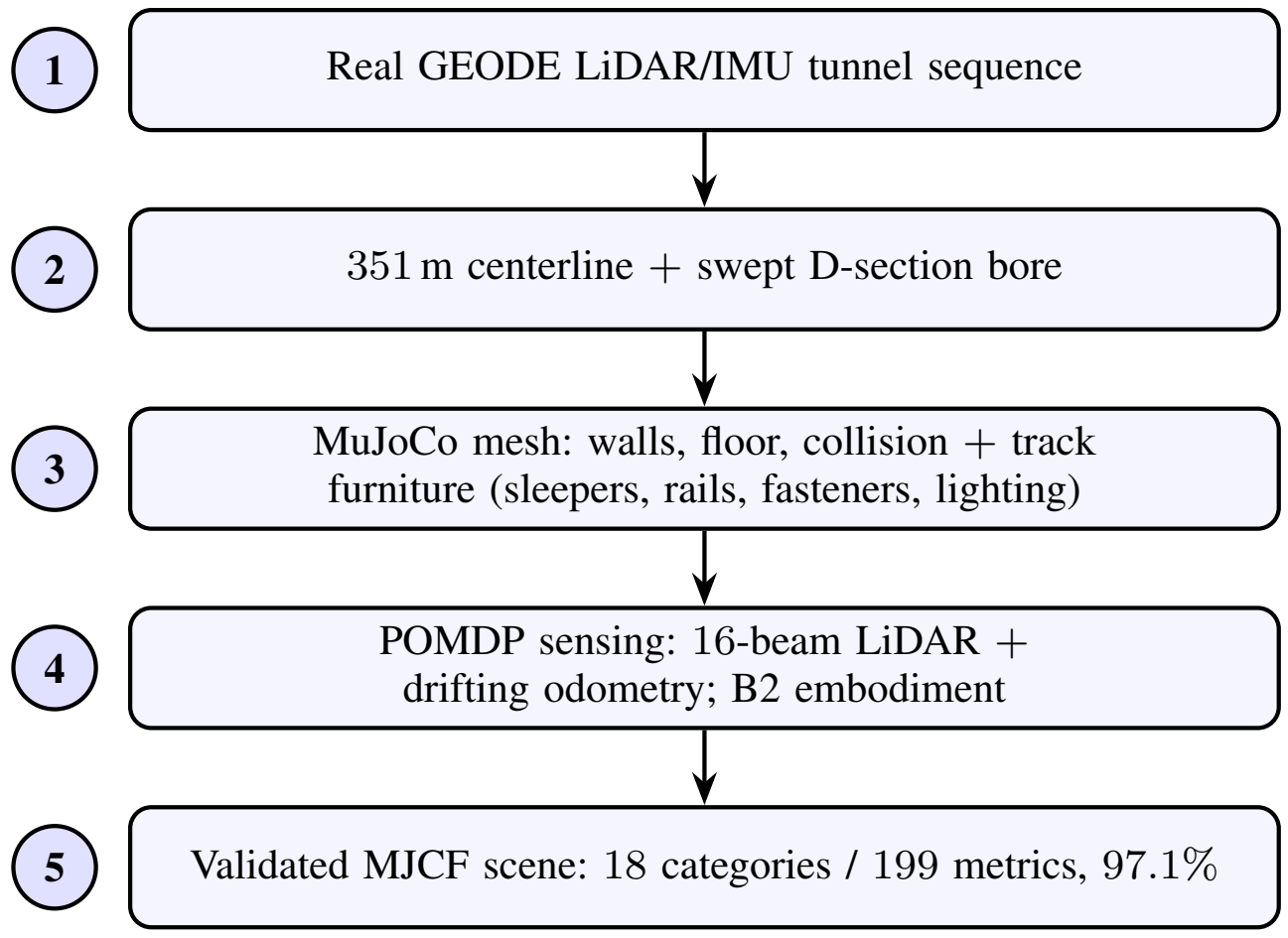
\begin{figure}[t]
\centering
\begin{tikzpicture}[
  node distance=5mm and 6mm,
  stage/.style={rectangle, rounded corners, draw, thick, fill=blue!4,
    text width=0.84\columnwidth, align=center, minimum height=8mm,
    inner sep=3pt, font=\footnotesize},
  num/.style={circle, draw, thick, fill=blue!12, minimum size=5.5mm,
    inner sep=0pt, font=\footnotesize\bfseries},
  arr/.style={-{Stealth[length=2.2mm]}, thick}]
\node[stage] (s1) {Real GEODE LiDAR/IMU tunnel sequence};
\node[stage, below=of s1] (s2) {$351$\,m centerline $+$ swept D-section bore};
\node[stage, below=of s2] (s3) {MuJoCo mesh: walls, floor, collision $+$ track furniture (sleepers, rails, fasteners, lighting)};
\node[stage, below=of s3] (s4) {POMDP sensing: $16$-beam LiDAR $+$ drifting odometry; B2 embodiment};
\node[stage, below=of s4] (s5) {Validated MJCF scene: $18$ categories / $199$ metrics, $97.1\%$};
\foreach \i in {1,...,5}{\node[num, left=of s\i.west, xshift=4mm] {\i};}
\draw[arr] (s1) -- (s2);
\draw[arr] (s2) -- (s3);
\draw[arr] (s3) -- (s4);
\draw[arr] (s4) -- (s5);
\end{tikzpicture}
\caption{Subterra construction pipeline: from
the raw GEODE recording to the released, validated MJCF scene.}
\label{fig:pipeline}
\end{figure}

\section{Sim-vs-Real Fidelity Validation}
\label{sec:val}
Subterra is validated quantitatively with a reproducible suite that scores the
reconstruction across $18$ independent categories---spanning geometry, mesh
integrity, topology, 2-D/3-D free space, navigability, point-cloud agreement,
occupancy, and scene assets---emitting $199$ logged metrics; each category
reduces to a percentage and the global score is their unweighted mean. The full
suite is reproducible from a single config (MuJoCo~$3.8.1$, fixed RNG seed) and is
shown insensitive to a $\pm10\%$ threshold sweep and to the seed, scoring a global
$\mathbf{97.1\%}$.

\begin{figure}[t]
\centering
\begin{subfigure}{\columnwidth}
\centering
\includegraphics[width=0.97\columnwidth,height=0.355\columnwidth, trim={0 0 0 60},clip]{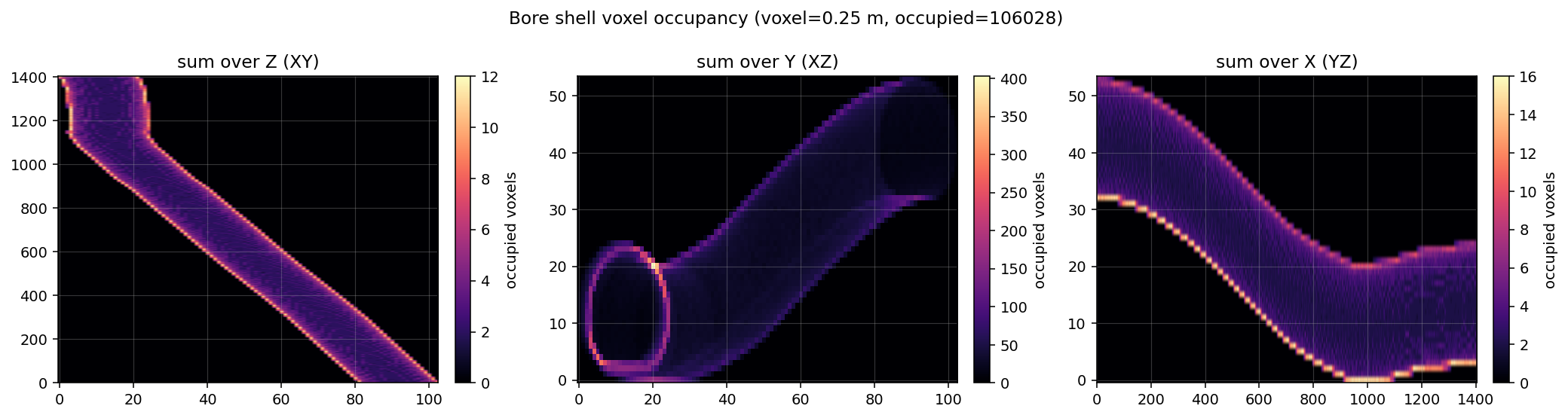}
\caption{Bore-shell voxel occupancy of the reconstruction, summed along each
axis (X--Y, X--Z, Y--Z).}
\label{fig:val_voxel}
\end{subfigure}\\[5pt]
\begin{subfigure}{\columnwidth}
\centering
\includegraphics[width=0.97\columnwidth,height=0.355\columnwidth]{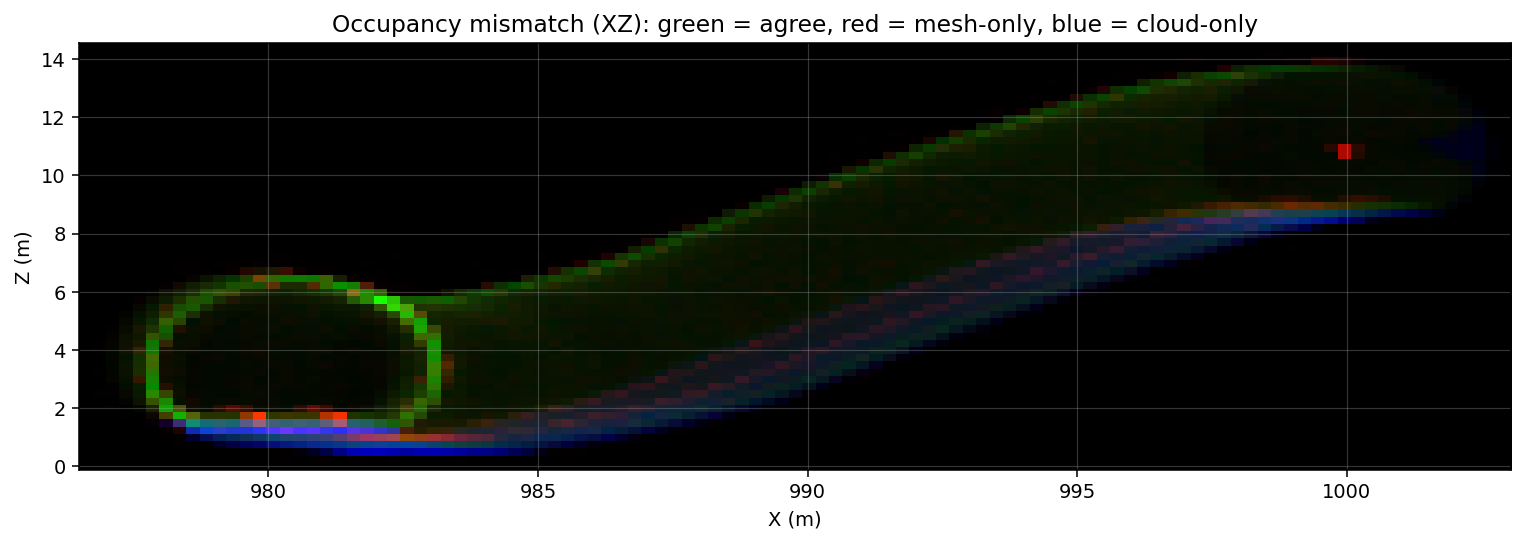}
\caption{Occupancy mismatch (X--Z): green$=$mesh and cloud agree,
red$=$mesh-only, blue$=$cloud-only.}
\label{fig:val_mismatch}
\end{subfigure}
\caption{Sim-vs-real spatial fidelity.}
\label{fig:val}
\end{figure}

The headline geometric result is an in-frame synthetic-LiDAR cloud that agrees
with the mesh to a mean point-to-surface error of $\mathbf{2.46}$\,\textbf{cm}
($97.5\%$ of points within $10$\,cm). Figure~\ref{fig:val} shows the spatial
fidelity directly: (a) the reconstructed bore's voxel occupancy and (b) a
voxel-by-voxel occupancy comparison against the real LiDAR cloud, in which
agreement (green) dominates and the few disagreements lie on thin boundary voxels
and the open bore mouth, not the tunnel structure.

\section{Learnability Check}
\label{sec:results}
\begin{figure}[t]
\centering
\includegraphics[width=0.98\columnwidth]{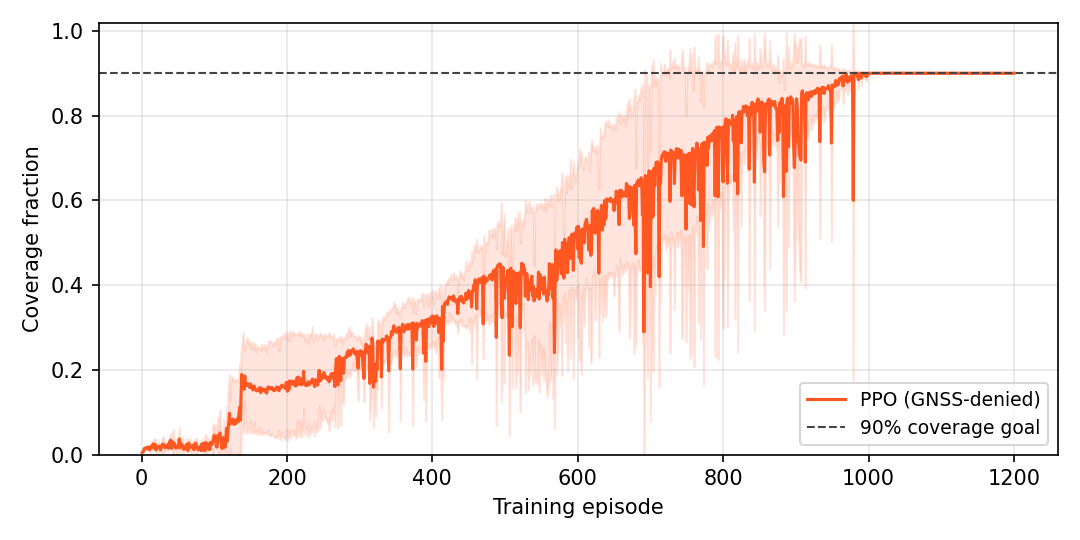}
\caption{Per-episode tunnel coverage. All seeds climb to and hold the $90\%$ coverage goal (dashed), confirming
the environment is learnable under partial observability.}
\label{fig:learn}
\end{figure}

We train a PPO
agent~\cite{schulman2017ppo} (Stable-Baselines3~\cite{raffin2021sb3}, frame-stacked
MLP policy) end-to-end on the partial observation, with no access to absolute
pose. The task is hard for an authentic reason: the bore is reconstructed from
real degenerate geometry, and the agent must close the loop through a noisy LiDAR
and drifting odometry rather than a tuned difficulty knob. We run five
independent seeds ($0$--$4$); coverage is reported as the mean$\pm$std across
seeds (Fig.~\ref{fig:learn}). All five seeds converge to the $90\%$ coverage goal
(mean rolling-$100$ coverage $90.02\%\pm0.01\%$, $100\%$ success at convergence), but only
after $\sim\!1000$ episodes of exploration---hard yet learnable, exactly the
regime a useful benchmark should occupy.

\section{Conclusion}
\label{sec:conclusion}
We presented \textbf{Subterra}, an open-source benchmark for GNSS-denied tunnel
navigation reconstructed from real GEODE LiDAR. We validated the reconstruction to
a global $97.1\%$ fidelity across $18$ categories---a $2.46$\,cm surface error
against the real reference---and showed that a PPO agent
learns to navigate it under partial observability, reaching the $90\%$ coverage
goal on all five seeds (mean rolling coverage $90.02\%\pm0.01\%$). Subterra, its construction pipeline, and the full
validation suite are released \textbf{open-source} as a faithful, ready-to-use
testbed for navigation under partial observability.


\bibliographystyle{plainnat}
\bibliography{references}

\end{document}